\RequirePackage{fix-cm}
\documentclass[smallcondensed]{svjour3}     
\smartqed  

\usepackage{epstopdf}
\usepackage{lscape}
\usepackage{filecontents}
\usepackage{adjustbox}
\usepackage{url}

\usepackage[font=small,labelfont=bf]{caption}
\usepackage{algcompatible}
\usepackage{multirow}
\usepackage{hyperref}
\usepackage[english]{babel}
\usepackage{algorithm}
\usepackage[noend]{algpseudocode}

\usepackage{graphicx}

\usepackage{colortbl}
\usepackage{color}
\usepackage[numbers]{natbib}
\usepackage{chngcntr}
\usepackage{longtable}

\usepackage{rotating}
\usepackage{subfig}

\usepackage{amsmath,amsfonts,amssymb,amsthm}

\theoremstyle{plain}

\newtheorem{theorem1}{Theorem}
\begin{document}
	
	\title{Online learning using multiple times weight updating \thanks{(a) This is a post-peer-review, pre-copyedit version of an article published in arxiv. The final authenticated version is available online at: \url{https://arxiv.org/abs/1811.00178} }
	}
	
	
	\author{Charanjeet $\color{red}^*$   \thanks{$\color{red}^*$corresponding author}         \and
		Anuj~Sharma
	}
	
	
	\institute{Charanjeet \at
		Computer Science \& Applications\\
		Panjab University Chandigarh, INDIA\\
		\email{charan@pu.ac.in,charanjeetpu@gmail.com}           
		\and
		Anuj Sharma \at
		Computer Science \& Applications\\
		Panjab University Chandigarh, INDIA\\
		\email{anujs@pu.ac.in}\\
		Homepage: \url{https://sites.google.com/view/anujsharma/}
		}
	
	\date{Received: date / Accepted: date}

	\maketitle
\begin{abstract}
Online learning makes sequence of decisions with partial data arrival where next movement of data is unknown. In this paper, we have presented a new technique as multiple times weight updating that update the weight iteratively forsame instance. The proposed technique analyzed with popular state-of-art algorithms from literature and experimented using established tool. The results indicates that mistake rate reduces to zero or close to zero for various datasets and algorithms. The overhead running cost is not too expensive and achieving mistake rate close to zero further strengthen the proposed technique. The present work include bound nature of weight updating for single instance and achieve optimal weight value. This proposed work could be extended to big datasets problems to reduce mistake rate in online learning environment. Also, the proposed technique could be helpful to meet real life challenges.
\end{abstract}

\section{Introduction}
\label{intro}
The Machine learning is one of the solution to the real life problems. Online learning is sub-field of machine learning. Online learning includes mainly weight updation with respect to minimization of loss. The online learning overcome the batch based system limitations in the situations, where training of model with respect to partial data arrival or real time application with unknown next movement of data. We have witnessed efficient algorithms in online learning from year 2000 onwards.
These algorithms were regularly experimented with new data sets and it did help to explore new algorithms in online learning. We have presented selected literature in online learning and the most of the techniques discussed in literature have been used with proposed method in experimentation.
One of early online learning algorithm was Perceptron \cite{Percept1957}. It is inspired by the information processing of neural cells called a neuron. The prediction of the perceptron algorithm based on a linear prediction function that combines a set of weighted vector and the training vector.
The Relaxed Online Maximum Margin Algorithm (ROMMA) \cite{ROMMA2002} is an incremental approach based on the maximum margin. ROMMA used the linear threshold function for classification. The maximum margin function can be formulated by minimizing the length of target vector subject to the number of linear constraint.
Approximate Large Margin Classification Algorithm (ALMA) \cite{ALMA2001} is an incremental algorithm, which approximate the maximal $p-norm$ margin for the set of linear separable data. ALMA works directly with the primal of the maximal margin.
Online Gradient Descent (OGD) \cite{OGD2003} for online convex functions, motivated from the infinitesimal gradient ascent and it deals with the Euclidean geometry. OGD is more general than expert setting that it can handle an arbitrary sequence of convex functions.
The other algorithm as Second Order Perceptron (SOP) \cite{SOP2005} used the second order properties of the data for learning the linear threshold function, defined as an interaction between eigenvalues of the correlation matrix of the data and target vector. The performance analysis of SOP remains within the mistake bound model of the online learning. The mistake bound depends on the parameter controlling the sensitivity of the algorithm to the distribution of these eigenvalues.
The online Passive Aggressive (PA) \cite{PA2006} follow the margin based online learning. The learning strategy of PA is based on the loss function (Hinge loss). The updation is passive when the loss function value is zero otherwise aggressively update the classifier when the loss is non-zero. PA updates classifier in such a manner that new update classifier should stay as close as to the previous classifier. It fails when the incoming data is non separable. To overcome above limitation there are two variant of PA. PAI and PAII balance the trade-off between the "passiveness" and "aggressiveness" using the positive parameter C called the aggressive parameter.
Online Newton Step (ONS) \cite{ONS2007} algorithm, which achieve the logarithmic loss for any arbitrary sequence of strictly convex functions. ONS use the second order information of the loss function and is based upon newton method for offline classification. ONS show the connection between follow-the-leader and Newton Method. It provides a logarithmic  regret for higher order derivative.
The Confidence-Weighted linear classification (CW) \cite{CW2008} algorithm is defined over the notion of confidence parameter. The less confident parameters are updated more aggressively than more confident ones. The confidence parameter is expressed in the term of Gaussian distribution over training vector.
Narrow Adaptive Regularization of Weights (NAROW) \cite{NAROW2010} allow to design and relative mistake bound for any loss function. The mistake bound for any loss function, allowing to recover and improve the bounds of online classification algorithms. The new online classification algorithm for optimize the general bound called NAROW, which makes use of adaptive and fixed-second order information. NAROW also provide bound for diagonal matrices.
A new algorithm based upon the velocity constraint in an online learning algorithm. In the learning process of Normal Herd (NHERD) \cite{NHERD2010} regularization of linear velocity term are used to herd the normal distribution. NHERD update is more aggressive for diagonal covariance matrix.
Double Updating Online Learning algorithm (DUOL) \cite{DUOL2011} is other online learning algorithm, when incoming instance is misclassified, it will be added into the pool of support vector and assigned with a weight, which often remain unchanged during the rest of the learning process. DUOL is dynamically tune the weights of the support vector in order to improve the classification performance.
This is an online learning technique that perform better in the presence of noisy label data. The Adaptive Regularization of Weight vector (AROW) \cite{AROW2009} is variant of confidence weight learning, beside that it holds various desirable properties of online learning algorithms: (1) confidence weighting, (2) large margin training and (3) handle the non-separable data.
LIBOL is an open-source library for large-scale online learning algorithms \cite{LIBOL2014} which includes all the state-of-arts algorithms for online classification.
SOLAR (Scalable Online Learning Algorithms for Ranking) \cite{SOLAR2015} learning to rank is learn some ranking model from training data using machine learning method, which is a type of information retrieval. This algorithm learn a ranking model from sequence of training data in an online learning fashion. This algorithm tackle the pairwise learning to ranking problem using scaleable online learning approach.
Soft Confidence-Weighted Learning (SCW) \cite{SCW2016}, which is the variant of confidence-weighted (CW) capable to handle non-separable cases, that is the limitation of CW. It is first online learning algorithm that hold the four silent properties: (1) confidence weighting, (2) capable to handle non-separable data, (3) large margin training, (4) adaptive margin. SCW exploits the adaptive margin by assigning different margin to different vector via a probability formulation.
Online Bayesian Passive Aggressive (BayesPA) (\cite{OBays2017}) framework for Bayesian models with maximum margin posterior regularization. For great flexibility and explorative analysis, BayesPA perform non-parametric Bayesian inference.
A survey on online learning algorithms \cite{SOL2018}, which presents state-of-art algorithms in this research field and their behavior has been discussed recently. It includes categorization of the online learning in three types: (1) Online supervised learning (2) online learning with limited feedback (3) Online unsupervised learning.\\
To the best of our knowledge, we are first to introduce MTWU model updating using multiple iteration for the same data points and our finding proves the efficiency of MTWU. The MTWU applied to all popular online learning algorithms including binary and multiclass environment with benchmark data sets. Our method establish the fact that most of the online learning algorithms reduce mistake rate to very low value. This paper include four section including this section as introduction. The section 2 present preliminaries of online learning and proposed method has been discussed in section 3. The section 4 presents the experimentation using benchmark data sets and future scope of present work. \\

\section{Preliminaries}
\label{prelim}
This section include the working of online learning algorithms that handle the data points in the form $(x_i,y_i)$ where $y_i$ is the class label of instance $x_i$. The online algorithm works in rounds where $x_i$ and its prediction function is $h(x_i)$. The prediction results is class label $\hat{y_i}$ and the loss function is $L(y_i,\hat{y_i})$. This update the model with prediction rule $h$ and form problems to minimize the                 loss as $\sum_{i=1}^{n} L(y_i,\hat{y_i}).$ the Algorithm 1 presents the nature of simple online learning algorithm. \\
\begin{algorithm}
\caption{Working of Online Learning Algorithm}\label{euclid}
\begin{algorithmic}[1]
\State Initialize $w_i=0$
\For{i=1 to n} \Comment{n is number of data point}
\State Predict $\hat{y_i}=<w_i,x_i>$
\State Compute Loss as $L(y_i,\hat{y_i})$
\If{$L(y_i,\hat{y_i}) > 0$}
\State $w_{i+1}= w_i+ <update \ rule> $ \Comment{Updating rule is depend on the selected algorithm}
\EndIf \label{euclideanIf}
\EndFor\label{euclidendfor}
\end{algorithmic}
\end{algorithm}\\
The goal is to minimize the loss value, which is used in predication task in learning method. It takes target value as input and determined the loss i.e difference between target value and the predicated value.
Few common types of loss functions are hinge loss and squared error loss \\
For the "Maximum Margin" classification hinge loss is the most promising function. For the predicted value $\hat{y_i}$ is defined as:
\begin{equation} \label{hinge}
L(y)=max(0,1-y_i.\hat{y_i})
\end{equation}
Note that $\hat{y_i}$ output of the classifier function.\\
Quadratic loss is also called Mean Square Error (MSE), which is commonly used for regression loss functions. Quadratic loss is the sum of squared difference between the actual output and the predicated output.\\
\begin{equation}
L(y)=\frac{\sum_{i=1}^n (y_i-\hat{y_i})}{n}
\end{equation}
Convergence of logistic loss and hinge loss is similar, but logistic is continues. The continues property of logistic loss may be utilized by the gradient descent method. At any point logistic loss does not assign a zero penalty.
\begin{equation}
L(y)=log(1+e^{-y_t   . \hat{y}_t})
\end{equation}
The update rule values vary with respect different algorithms. For example, few selected update rules are discussed in following paragraph.\\
The loss function use by PA (\cite{PA2006}) is in equation \ref{hinge}. The updation is passive when $l=0$ otherwise aggressively updation come into an action. The closed form updation rules of three variant of PA is
\begin{equation} \label{equPA}
w_{t+1}=w_t + \tau_t y_tx_t \ , \ \tau_t = \begin{cases}
 l_t/||x_t||^2 , & \text{PA}\\
 min ( C, l_t/||x_t||^2 ), &  \text{PA1}\\
 \frac{l_t}{||x_t||^2 + \frac{1}{2C}},&  \text{PA2  }
\end{cases}
\end{equation}

The OGD \cite{OGD2003} used to solve the online convex optimization problems. The OGD used equation \ref{hinge} as a loss function and updation rule is:
\begin{equation}
w_{t+1}= w_t + \eta_ty_tx_t
\end{equation}
OGD use some predefined learning rate ($\eta_t$). \\
SOP \cite{SOP2005} is the incremental variant of whitened perceptron algorithm. The weight updation strategy of SOP is: \\
\begin{equation}
v_k=v_{k-1}+ y_tx_t \ ,
X_k=S_t
\end{equation}
The SOP predication is computed in trial $t$, use  $v_{k-1}$ an $n$-dimensional weight vector and $X_{k-1}$ use $n$-row matrix, where subscript $k-1$ indicates the number of times vector $v$ and the matrix $X$ have been updated in the first $t-1$ trials.\\
ONS \cite{ONS2007} is the online variant of the Newton-Raphson method and use the second order properties of the loss function. The updation rule of ONS is:\\
\begin{equation}
x_t=\Pi_p^{A_{t-1}} (x_{t-1}-\frac{1}{\beta}A_{t-1}^{-1} \nabla_{t-1})
\end{equation}
Where $\nabla_t$ and $A_t$ are gradient and hessian values. In this algorithm projection is according to the norm defined by the matrix $A_t$.
CW \cite{CW2008} learning method for linear classification is based upon standard deviation. CW update the weight that is based upon the confidence of the weight vector. The confidence of the weight vector is calculated using the Gaussian distribution and the covariance matrix. The updation rule of CW is: \\
\begin{equation}
( \mu_{t+1},\Sigma_{t+1} ) = arg \ min \ D_{KL} (N(\mu, \Sigma) || (\mu_{t}, \Sigma_{t}))
\end{equation}
$\mu$ is the mean vector and $\Sigma$ is covariance matrix. $D_{KL}$ is the KL divergence distance between two distributions.
The online algorithms are successfully applied to binary and multiclass data. In literature, all the successful online learning algorithms have been proved with upper bound mistake rate that further prove the strong mathematical foundations behind these techniques.

\section{The MTWU Step}
\label{MTWU}
Our proposed MTWU is applicable to mostly state-of-art algorithms. The MTWU include simple but powerful step as multiple times weight updating of single instance. The algorithm 2 presents the working of MTWU.
\begin{algorithm}
\caption{MTWU }\label{euclid}
\begin{algorithmic}[1]
\State Initialize $w_i=0$
\For{i=1 to $n$} \Comment{n is the number of data point}
\For {k=1 to $m$} \Comment{where m=1 to n}
\State Predict $\hat{y_i}=<w_i,x_i>$
\State Compute Loss as $L(y_i,\hat{y_i})$
\If{$L(y_i,\hat{y_i}) > 0$}
\State $w_{i+1}= w_i+ <update \ rule> $ \Comment{Updating rule depends on the \hspace{10ex} selected algorithm}

\EndIf \label{euclideanIf}
\EndFor
\EndFor\label{euclidendfor}
\end{algorithmic}
\end{algorithm}\\
A loop is applied to train the weights for one instance at a time that results in loss minimization and weights are trained optimally. We have noticed less mistake rate at $m=2$ and achieve constant mistake rate (zero in some cases) from $m=8$ onwards. The updation for mostly cases improve $m=2$ onwards where mistake rate appears zero for few data sets. The MTWU do not include any other changes in established algorithms other than introduction of loop. Also, no changes are made to feature vector and predicted class in each iteration of introduced loop. The weights are updated in each iteration subject to the dependent values used in the each algorithm and in single iteration of respective algorithm. The result of MTWU are discussed in next section to prove the efficiency of proposed method. \\
As MTWU used with established online learning techniques, we noticed that algorithm used with MTWU has been discussed in literature thoroughly. This also includes regret bound for respective algorithms. Our MTWU is a step that repeat definite number of times, therefore, it do not interfere with regret bound of used algorithms. We have presented Theorem 1 to establish that for a single instance to achieve optimum value is bounded. \\
\begin{theorem1}
	\label{theorem}
The weight $w_i$ at $i^{th}$ iteration updated multiple times achieve optimum value $w_i^*$ bounded as:
\begin{equation}
0 \leq ||w_i^*|| \leq ||w_i^0|| + \sqrt{M} ( \sum_{j=1}^{M} ||\Delta w_i^j||^2)^\frac{1}{2}
\end{equation}
\end{theorem1}
\begin{proof}
 Let $w_i$ is the $i^{th}$ data point update for $n$ data points and $w_i^k$ is the $i^{th}$ data point updated $k$ number of points.
Let $\Delta w_i$ is update rule value for respective $i^{th}$ data point and $\Delta w_i^k$ is the update rule value for $k^{th}$ iteration of $i^{th}$ data point. A weight update is,
\begin{equation*}
w_i=w_{i-1}+ \Delta w_i
\end{equation*}

and weight update for $k^{th}$ iteration at $i^{th}$ point is,
\begin{equation*}
w_i^k=w_{i-1}^{k}+ \Delta w_i^k
\end{equation*}
Let $w_i^*$ is the optimum weight at $w_i^k$, therefore,
\begin{equation*}
\begin{array}{l}
 w_{i}^{*}=w_{i}^{k-1} + \Delta w_i^k \\
 \hspace{0.5cm} =w_{i}^{k-2}+ \Delta w_{i}^{k-1}+ \Delta w_i^k  \\
 \hspace{0.5cm}=w_i^0+\Delta w_i^1+ \Delta w_i^2 + .... + \Delta w_i^k
 \end{array}
 \end{equation*}
 \begin{equation*}
 w_i^*-w_i^0 = \Delta w_i^1+ \Delta w_i^2 + .... + \Delta w_i^k
 \end{equation*}
 using norm and square both sides,
 \begin{equation*}
 \begin{array}{llll}
||w_i^*-w_i^0||^2 = ||\Delta w_i^1+ \Delta w_i^2 + .... + \Delta w_i^k||^2  \\
 ||w_i^*-w_i^0||^2 = ||\Delta w_i^1 \times 1+ \Delta w_i^2 \times 1 + .... + \Delta w_i^k \times 1||^2
 \end{array}
 \end{equation*}
 using cauchy schwartz inequality,\\
 \begin{equation*}
 (\sum_{i=1}^{n}a_i b_i)^2 \leq (\sum_{i=1}^{n}a_i^2) (\sum_{i=1}^{n}b_i^2)
 \end{equation*}
 \begin{equation*}
 \begin{array}{ll}
 ||w_i^*-w_i^0||^2 \leq (|| \Delta w_i^1||^2 + || \Delta w_i^2||^2 +..... + ||\Delta w_i^k||^2) \\
\hspace{1.9cm} (1^2+1^2+..... +1^2)\\
 \hspace{1.9cm} \leq  (\sum_{j=1}^{K} ||\Delta w_i||^2)\times K , \hspace{1.8cm} \textit{let M $\geq$ K} \\
  \hspace{1.9cm} \leq M (\sum_{j=1}^{M} ||\Delta w_i||^2)
   \end{array}
   \end{equation*}
   \begin{equation*}
   \begin{array}{lll}
  ||w_i^*-w_i^0|| \leq \sqrt{M} (\sum_{j=1}^{M} ||\Delta w_i^j||^2)^\frac{1}{2} \\
  using \hspace{1cm}||w_i^*-w_i^0|| \geq ||w_i^*||-||w_i^0|| \geq -||w_i^0||
  \end{array}
  \end{equation*}

\begin{equation*}
 -||w_i^0|| \leq ||w_i^*||-||w_i^0|| \leq \sqrt{M} (\sum_{j=1}^{M} ||\Delta w_i^j||^2)^\frac{1}{2}
 \end{equation*}

  adding $||w_i^0||$ above
  \begin{equation*}
  0 \leq ||w_i^*|| \leq ||w_i^0||+ \sqrt{M} (\sum_{j=1}^{M} ||\Delta w_i^j||^2)^\frac{1}{2}
  \end{equation*}
Hence it proves the result.
\end{proof}

\section{Experimental results}
\label{experi}
In this section, we apply MTWU to popular and selected online learning algorithms mentioned in section 1 introduction. The benchmark datasets are used and experiments are conducted for both binary and multiple classes datasets. We have used benchmark tool as Libol (\cite{LIBOL2014}) to prove the effectiveness of our proposed technique MTWU. The Table 1 present names for online learning algorithms as used in tool Libol. The experiments are performed in machine with i7 processor and 8GB ram.\\

\begin{table}[ht]
\caption {Online learning algorithms and their used abbreviations} \label{tab:tab1}
\begin{adjustbox}{width=1\textwidth}
\begin{tabular}{|l|l|l|}
\hline
\textbf{ Algorithm Name  }                                                                            & \textbf{Name }     & \textbf{References  }              \\ \hline
The Second Order Perceptron (SOP) algorithm                                                 & SOP        & Cesa-Bianchi et al., \cite{SOP2005} \\ \hline
\begin{tabular}[c]{@{}l@{}}The Confidence-Weighted (CW)\\   learning algorithm\end{tabular} & CW         & Dredze et al.,       \cite{CW2008} \\ \hline
Online learning algorithms by improved ellipsoid method                                     & IELLIP     & Yang et al.,         \cite{IEELIP2009} \\ \hline
The Adaptive Regularization of Weight Vectors                                               & AROW       & Crammer et al.,      \cite{AROW2009} \\ \hline
New variant of Adaptive Regularization                                                      & NAROW      & Orabona and Crammer, \cite{NAROW2010} \\ \hline
The Normal Herding method via Gaussian Herding                                              & NHERD      & Crammer and Lee,    \cite{NHERD2010} \\ \hline
The recently proposed Soft Confidence Weighted algorithms                                   & SCW        & Wang et al.,       \cite{SCW2016} \\ \hline
the classical online learning algorithm                                                     & Perceptron & Rosenblatt,         \cite{Percept1957} \\ \hline
A New Approximate Maximal Margin Classification Algorithm                                   & ALMA       & Gentile,            \cite{ALMA2001}\\ \hline
The relaxed online maximum margin algorithms                                                & ROMMA      & Li and Long,         \cite{ROMMA2002} \\ \hline
The Online Gradient Descent (OGD) algorithms                                                & OGD        & Zinkevich,          \cite{OGD2003} \\ \hline
the Passive Aggressive (PA) online learning algorithms                                      & PA         & Crammer et al.,      \cite{PA2006} \\ \hline
\end{tabular}
\end{adjustbox}
\end{table}

\subsection{Binary class datasets}
The binary datasets used are svmguide3 and covtype. The svmguide3 includes 1243 data points 21 features. we have used MTWU for $m=1,2,4,8,16,32$ for each algorithm. This $m$ is the variable used for iteration in algorithm \ref{euclid} at line 3. Table 2 present results for dataset svmguide3. We find that each algorithm in Table 2 achieve mistake rate as zero for some value of $m$. The algorithms SOP, \ SCW,\ PA2, \ PA1, \ OGD, \ CW, \ ALMA and IEELIP achieve zero mistake rates at $m=4$ where as algorithm SCW2 at $m=8$. Table 3 present covtype dataset results that achieve mistake rate as zero using the algorithms PA2, \ PA1, \ PA, \ ALMA, \ aROMMA, \ and IELLIP at $m=2$ where as SOP and OGD at $m=8$. \\
\subsection{Multi class datasets}
The used datasets for mutiple classes are mnist, glass and segment. The dataset mnist include 60K data points and 780 feature in each data point for 10 classes. The segment dataset contains 2310 data points and 19 features for 7 classes respectively. The glass dataset include 214 data points and 300 features for 6 classes. Similar to the binary class experiments, MTWU applied with $m=1,2,4,8,16$ and $32$. Table 4 presents mnist dataset results that achieve mistake rate as zero at some value of $m$. The algorithm M\_PA, M\_PA1, M\_PA2, M\_ROMMA, M\_PerceptronS, M\_PerceptronM, M\_PerceptronU, M\_SCW2 and M\_CW achieves zero mistake at $m=4$ where as algorithm M\_OGD achieve mistake rate zero at $m=8$.
 Table 5 presents glass dataset results where algorithms M\_PA, M\_PA1, M\_PA2,  M\_PerceptronS, M\_PerceptronM, M\_PerceptronU, M\_SCW2 and M\_CW achieves zero mistake at $m=4$ where as algorithm M\_OGD, M\_ROMMA, M\_aROMMA achieve mistake rate zero at $m=8$.
   Table 6 presents segment dataset results, the algorithm M\_PA, M\_PA1, M\_PA2,  M\_PerceptronS, M\_PerceptronM, M\_PerceptronU, M\_SCW2 and M\_CW achieve zero mistake at $m=4$ where as algorithm M\_OGD, M\_ROMMA, M\_aROMMA achieve mistake rate zero at $m=8$.
\begin{landscape}
\begin{table}[ht]
\caption {Results for svmguide3 binaryclass dataset} \label{tab:tab2}
\begin{adjustbox}{width=1.5\textwidth}
\begin{tabular}{|l|l|l|l|l|l|l|l|}
\hline
\multicolumn{8}{|l|}{\textbf{Dataset name: svmguide3 (n=1243,d=21,No. of classes=2)    nb of runs (permutations): 20}}                                                                    \\ \hline
\textbf{Algorithm}                   &               & \textbf{m=1}               &\textbf{ m=2}               & \textbf{m=4}               & \textbf{m=8}               & \textbf{m=16}              & \textbf{m=32}              \\ \hline
\multirow{3}{*}{SOP}        & Mistake Rate  & 0.3008 +/- 0.0098 & 0.0597 +/- 0.0067 & 0.0010 +/- 0.0009 & 0.0000 +/- 0.0000 & 0.0000 +/- 0.0000 & 0.0000 +/- 0.0000 \\ \cline{2-8}
                            & NB of Updates & 373.95 +/- 12.15  & 74.25 +/- 8.35    & 1.30 +/- 1.13     & 0.00 +/- 0.00     & 0.00 +/- 0.00     & 0.00 +/- 0.00     \\ \cline{2-8}
                            & Cpu Time      & 0.0981 +/- 0.0033 & 0.1198 +/- 0.0065 & 0.1579 +/- 0.0055 & 0.2330 +/- 0.0069 & 0.3820 +/- 0.0065 & 0.6900 +/- 0.0133 \\ \hline
\multirow{3}{*}{SCW}        & Mistake Rate  & 0.2096 +/- 0.0053 & 0.0000 +/- 0.0000 & 0.0000 +/- 0.0000 & 0.0000 +/- 0.0000 & 0.0000 +/- 0.0000 & 0.0000 +/- 0.0000 \\ \cline{2-8}
                            & NB of Updates & 590.95 +/- 11.89  & 320.75 +/- 17.55  & 35.90 +/- 5.12    & 12.70 +/- 3.20    & 8.95 +/- 2.87     & 16.85 +/- 3.42    \\ \cline{2-8}
                            & Cpu Time      & 0.0879 +/- 0.0006 & 0.1028 +/- 0.0010 & 0.1179 +/- 0.0010 & 0.1504 +/- 0.0013 & 0.2135 +/- 0.0012 & 0.3445 +/- 0.0027 \\ \hline
\multirow{3}{*}{ROMMA}      & Mistake Rate  & 0.3359 +/- 0.0146 & 0.3359 +/- 0.0146 & 0.3359 +/- 0.0146 & 0.3359 +/- 0.0146 & 0.3359 +/- 0.0146 & 0.3359 +/- 0.0146 \\ \cline{2-8}
                            & NB of Updates & 417.55 +/- 18.13  & 417.55 +/- 18.13  & 417.55 +/- 18.13  & 417.55 +/- 18.13  & 417.55 +/- 18.13  & 417.55 +/- 18.13  \\ \cline{2-8}
                            & Cpu Time      & 0.0670 +/- 0.0033 & 0.0764 +/- 0.0052 & 0.0876 +/- 0.0043 & 0.1117 +/- 0.0042 & 0.1562 +/- 0.0052 & 0.2484 +/- 0.0054 \\ \hline
\multirow{3}{*}{Perceptron} & Mistake Rate  & 0.3304 +/- 0.0101 & 0.3304 +/- 0.0101 & 0.3304 +/- 0.0101 & 0.3304 +/- 0.0101 & 0.3304 +/- 0.0101 & 0.3304 +/- 0.0101 \\ \cline{2-8}
                            & NB of Updates & 410.65 +/- 12.51  & 410.65 +/- 12.51  & 410.65 +/- 12.51  & 410.65 +/- 12.51  & 410.65 +/- 12.51  & 410.65 +/- 12.51  \\ \cline{2-8}
                            & Cpu Time      & 0.0704 +/- 0.0109 & 0.0685 +/- 0.0027 & 0.0777 +/- 0.0035 & 0.0919 +/- 0.0036 & 0.1181 +/- 0.0039 & 0.1680 +/- 0.0041 \\ \hline
\multirow{3}{*}{PA2}        & Mistake Rate  & 0.2551 +/- 0.0061 & 0.0000 +/- 0.0000 & 0.0000 +/- 0.0000 & 704.85 +/- 13.59  & 0.0000 +/- 0.0000 & 0.0000 +/- 0.0000 \\ \cline{2-8}
                            & NB of Updates & 1135.10 +/- 12.29 & 724.85 +/- 14.57  & 724.70 +/- 14.65  & 704.85 +/- 13.59  & 26.15 +/- 8.65    & 20.40 +/- 6.37    \\ \cline{2-8}
                            & Cpu Time      & 0.0732 +/- 0.0018 & 0.0749 +/- 0.0013 & 0.0820 +/- 0.0007 & 0.0970 +/- 0.0007 & 0.1167 +/- 0.0025 & 0.1558 +/- 0.0023 \\ \hline
\multirow{3}{*}{PA1}        & Mistake Rate  & 0.2369 +/- 0.0016 & 0.0000 +/- 0.0000 & 0.0000 +/- 0.0000 & 0.0000 +/- 0.0000 & 0.0000 +/- 0.0000 & 0.0000 +/- 0.0000 \\ \cline{2-8}
                            & NB of Updates & 768.35 +/- 11.78  & 261.75 +/- 14.56  & 261.75 +/- 14.56  & 20.80 +/- 6.78    & 15.70 +/- 6.26    & 13.65 +/- 6.25    \\ \cline{2-8}
                            & Cpu Time      & 0.0692 +/- 0.0006 & 0.0730 +/- 0.0033 & 0.0730 +/- 0.0033 & 0.0856 +/- 0.0027 & 0.1041 +/- 0.0010 & 0.1447 +/- 0.0063 \\ \hline
\multirow{3}{*}{PA}         & Mistake Rate  & 0.3241 +/- 0.0086 & 0.3241 +/- 0.0086 & 0.3241 +/- 0.0086 & 0.3241 +/- 0.0086 & 0.3241 +/- 0.0086 & 0.3241 +/- 0.0086 \\ \cline{2-8}
                            & NB of Updates & 724.70 +/- 14.65  & 724.70 +/- 14.65  & 724.70 +/- 14.65  & 724.70 +/- 14.65  & 724.70 +/- 14.65  & 724.70 +/- 14.65  \\ \cline{2-8}
                            & Cpu Time      & 0.0688 +/- 0.0072 & 0.0747 +/- 0.0040 & 0.0877 +/- 0.0126 & 0.1036 +/- 0.0015 & 0.1450 +/- 0.0053 & 0.2210 +/- 0.0059 \\ \hline
\multirow{3}{*}{OGD}        & Mistake Rate  & 0.2377 +/- 0.0021 & 0.1253 +/- 0.0061 & 0.0091 +/- 0.0023 & 0.0000 +/- 0.0002 & 0.0000 +/- 0.0000 & 0.0000 +/- 0.0000 \\ \cline{2-8}
                            & NB of Updates & 637.55 +/- 3.20   & 261.05 +/- 8.57   & 32.65 +/- 5.55    & 0.45 +/- 0.60     & 0.00 +/- 0.00     & 0.00 +/- 0.00     \\ \cline{2-8}
                            & Cpu Time      & 0.0687 +/- 0.0006 & 0.0730 +/- 0.0015 & 0.0808 +/- 0.0006 & 0.0959 +/- 0.0007 & 0.1266 +/- 0.0011 & 0.1878 +/- 0.0023 \\ \hline
\multirow{3}{*}{NHERD}      & Mistake Rate  & 0.2279 +/- 0.0158 & 0.2085 +/- 0.0114 & 0.2032 +/- 0.0074 & 0.1998 +/- 0.0069 & 0.1969 +/- 0.0064 & 0.1949 +/- 0.0061 \\ \cline{2-8}
                            & NB of Updates & 1149.55 +/- 31.88 & 1153.25 +/- 26.83 & 1150.80 +/- 24.92 & 1148.10 +/- 25.39 & 1150.50 +/- 23.20 & 1139.05 +/- 25.33 \\ \cline{2-8}
                            & Cpu Time      & 0.0548 +/- 0.0013 & 0.0658 +/- 0.0010 & 0.0878 +/- 0.0013 & 0.1409 +/- 0.0057 & 0.2212 +/- 0.0045 & 0.3900 +/- 0.0079 \\ \hline
\multirow{3}{*}{NAROW}      & Mistake Rate  & 0.3007 +/- 0.0510 & 0.2870 +/- 0.0509 & 0.2870 +/- 0.0509 & 0.2870 +/- 0.0509 & 0.2870 +/- 0.0509 & 0.2870 +/- 0.0509 \\ \cline{2-8}
                            & NB of Updates & 1191.00 +/- 19.27 & 1191.00 +/- 19.27 & 1191.00 +/- 19.27 & 1191.00 +/- 19.27 & 1191.00 +/- 19.27 & 1191.00 +/- 19.27 \\ \cline{2-8}
                            & Cpu Time      & 0.0567 +/- 0.0006 & 0.0747 +/- 0.0019 & 0.1029 +/- 0.0011 & 0.1580 +/- 0.0047 & 0.2741 +/- 0.0043 & 0.4932 +/- 0.0091 \\ \hline
\multirow{3}{*}{CW}         & Mistake Rate  & 0.2912 +/- 0.0089 & 0.0000 +/- 0.0000 & 0.0000 +/- 0.0000 & 0.0000 +/- 0.0000 & 0.0000 +/- 0.0000 & 0.0000 +/- 0.0000 \\ \cline{2-8}
                            & NB of Updates & 700.70 +/- 18.28  & 357.50 +/- 21.76  & 78.95 +/- 9.65    & 32.45 +/- 5.41    & 25.85 +/- 6.34    & 24.50 +/- 4.86    \\ \cline{2-8}
                            & Cpu Time      & 0.0896 +/- 0.0009 & 0.1055 +/- 0.0039 & 0.1203 +/- 0.0013 & 0.1605 +/- 0.0110 & 0.2231 +/- 0.0034 & 0.3429 +/- 0.0020 \\ \hline
\multirow{3}{*}{AROW}       & Mistake Rate  & 0.2242 +/- 0.0034 & 0.1973 +/- 0.0058 & 0.1896 +/- 0.0047 & 0.1840 +/- 0.0051 & 0.1813 +/- 0.0056 & 0.1794 +/- 0.0047 \\ \cline{2-8}
                            & NB of Updates & 1221.05 +/- 6.49  & 1156.70 +/- 18.70 & 1149.00 +/- 20.45 & 1141.80 +/- 20.36 & 1137.60 +/- 20.93 & 1133.95 +/- 21.96 \\ \cline{2-8}
                            & Cpu Time      & 0.0992 +/- 0.0016 & 0.1183 +/- 0.0008 & 0.1569 +/- 0.0013 & 0.2321 +/- 0.0019 & 0.3844 +/- 0.0061 & 0.6730 +/- 0.0068 \\ \hline
\multirow{3}{*}{SCW2}       & Mistake Rate  & 0.2241 +/- 0.0148 & 0.0226 +/- 0.0033 & 0.0619 +/- 0.0037 & 0.0000 +/- 0.0000 & 0.0000 +/- 0.0000 & 0.0048 +/- 0.0021 \\ \cline{2-8}
                            & NB of Updates & 817.15 +/- 72.30  & 562.10 +/- 19.72  & 0.0619 +/- 0.0037 & 17.60 +/- 4.81    & 0.10 +/- 0.31     & 469.35 +/- 19.03  \\ \cline{2-8}
                            & Cpu Time      & 0.0960 +/- 0.0028 & 0.1084 +/- 0.0031 & 0.1508 +/- 0.0019 & 0.1624 +/- 0.0015 & 0.2251 +/- 0.0012 & 0.5452 +/- 0.0071 \\ \hline
\multirow{3}{*}{ALMA}       & Mistake Rate  & 0.2314 +/- 0.0043 & 0.0000 +/- 0.0000 & 0.0000 +/- 0.0000 & 0.0000 +/- 0.0000 & 0.0000 +/- 0.0000 & 0.0000 +/- 0.0000 \\ \cline{2-8}
                            & NB of Updates & 594.15 +/- 8.67   & 0.00 +/- 0.00     & 0.00 +/- 0.00     & 0.00 +/- 0.00     & 0.00 +/- 0.00     & 0.00 +/- 0.00     \\ \cline{2-8}
                            & Cpu Time      & 0.0772 +/- 0.0053 & 0.0729 +/- 0.0013 & 0.0787 +/- 0.0012 & 0.0923 +/- 0.0048 & 0.1122 +/- 0.0038 & 0.1538 +/- 0.0008 \\ \hline
\multirow{3}{*}{aROMMA}     & Mistake Rate  & 0.3315 +/- 0.0144 & 0.3315 +/- 0.0144 & 0.3315 +/- 0.0144 & 0.3315 +/- 0.0144 & 0.3315 +/- 0.0144 & 0.3315 +/- 0.0144 \\ \cline{2-8}
                            & NB of Updates & 1506.90 +/- 27.08 & 506.90 +/- 27.08  & 506.90 +/- 27.08  & 506.90 +/- 27.08  & 506.90 +/- 27.08  & 506.90 +/- 27.08  \\ \cline{2-8}
                            & Cpu Time      & 0.0721 +/- 0.0126 & 0.0783 +/- 0.0058 & 0.0944 +/- 0.0119 & 0.1194 +/- 0.0119 & 0.1692 +/- 0.0052 & 0.2711 +/- 0.0082 \\ \hline
\multirow{3}{*}{IELLIP}     & Mistake Rate  & 0.3359 +/- 0.0065 & 0.0000 +/- 0.0000 & 0.0000 +/- 0.0000 & 0.0000 +/- 0.0000 & 0.0000 +/- 0.0000 & 0.0000 +/- 0.0000 \\ \cline{2-8}
                            & NB of Updates & 417.50 +/- 8.09   & 0.00 +/- 0.00     & 0.00 +/- 0.00     & 0.00 +/- 0.00     & 0.00 +/- 0.00     & 0.00 +/- 0.00     \\ \cline{2-8}
                            & Cpu Time      & 0.0845 +/- 0.0006 & 0.0916 +/- 0.0018 & 0.1074 +/- 0.0011 & 0.1392 +/- 0.0015 & 0.1994 +/- 0.0007 & 0.3204 +/- 0.0017 \\ \hline
\end{tabular}
\end{adjustbox}
\end{table}
\end{landscape}

\begin{landscape}
\begin{table}[ht]
\caption {Results for covtype binary class dataset} \label{tab:tab3}
\begin{adjustbox}{width=1.5\textwidth}
\begin{tabular}{|l|l|l|l|l|l|l|l|}
\hline
\multicolumn{8}{|l|}{\textbf{Dataset name: covtype (n=581012,d=54,No. of classes=2)     nb of runs (permutations): 20} }                                                                                                                                                                                                                                \\ \hline
Algorithm                   &               & \textbf{m=1}                     & \textbf{m=2}                                                            &\textbf{ m=4  }                                                            & \textbf{m=8}                                                                & \textbf{m=16}                   & \textbf{m=32}                   \\ \hline
\multirow{3}{*}{SOP}        & Mistake Rate  & 0.3371 +/- 0.0005       & 0.0563 +/- 0.0003                                               & 0.0002 +/- 0.0000                                                & 0.0000 +/- 0.0000                                                  & 0.0000 +/- 0.0000      & 0.0000 +/- 0.0000      \\ \cline{2-8}
                            & NB of Updates & 195884.50 +/- 301.12    & 32700.20 +/- 165.14 & 141.00 +/- 13.10                                                 & 0.25 +/- 0.72                                                      & 0.00 +/- 0.00          & 0.00 +/- 0.00          \\ \cline{2-8}
                            & Cpu Time      & 26.7849 +/- 0.3193      & 41.5118 +/- 2.4081                                              & 53.7939 +/- 2.3029                                               & 92.0715 +/- 4.9958                                                 & 147.7129 +/- 8.2466    & 279.2301 +/- 8.3071    \\ \hline
\multirow{3}{*}{SCW}        & Mistake Rate  & 0.2328 +/- 0.0003       & 0.2227 +/- 0.0002                                               & 0.2190 +/- 0.0002                                                & 0.2173 +/- 0.0002                                                  & 0.2165 +/- 0.0002      & 0.2161 +/- 0.0002      \\ \cline{2-8}
                            & NB of Updates & 240297.70 +/- 2672.49   & 322288.15 +/- 163.88                                            & 308573.60 +/- 205.81                                             & 303179.90 +/- 221.76                                               & 300591.65 +/- 204.56   & 299165.85 +/- 213.22   \\ \cline{2-8}
                            & Cpu Time      & 24.8498 +/- 1.0065      & 31.6651 +/- 0.2514                                              & 42.7688 +/- 0.3963                                               & 69.2404 +/- 3.9498                                                 & 113.0031 +/- 4.1858    & 188.5614 +/- 0.6723    \\ \hline
\multirow{3}{*}{ROMMA}      & Mistake Rate  & 0.4833 +/- 0.0133       & 0.4833 +/- 0.0133                                               & 0.4833 +/- 0.0133                                                & 0.4833 +/- 0.0133                                                  & 0.4833 +/- 0.0133      & 0.4833 +/- 0.0133      \\ \cline{2-8}
                            & NB of Updates & 280789.90 +/- 7735.31   & 280789.90 +/- 7735.31                                           & 280789.90 +/- 7735.31                                            & 280789.90 +/- 7735.31                                              & 280789.90 +/- 7735.31  & 280789.90 +/- 7735.31  \\ \cline{2-8}
                            & Cpu Time      & 85.8472 +/- 6.2899      & 82.8738 +/- 2.1625                                              & 108.3522 +/- 24.9696                                             & 133.3594 +/- 3.7537                                                & 146.2412 +/- 4.1844    & 144.3720 +/- 3.5783    \\ \hline
\multirow{3}{*}{Perceptron} & Mistake Rate  & 0.4699 +/- 0.0004       & 0.4699 +/- 0.0004                                               & 0.4699 +/- 0.0004                                                & 0.4699 +/- 0.0004                                                  & 0.4699 +/- 0.0004      & 0.4699 +/- 0.0004      \\ \cline{2-8}
                            & NB of Updates & 273015.85 +/- 252.03    & 273015.85 +/- 252.03                                            & 273015.85 +/- 252.03& 273015.85 +/- 252.03                                               & 273015.85 +/- 252.03   & 273015.85 +/- 252.03   \\ \cline{2-8}
                            & Cpu Time      & 29.2189 +/- 0.2345      & 31.1626 +/- 0.5224                                              & 34.6718 +/- 0.3681                                               & 53.7110 +/- 4.5016                                                 & 57.1125 +/- 1.6677     & 273015.85 +/- 252.03   \\ \hline
\multirow{3}{*}{PA2}        & Mistake Rate  & 0.4833 +/- 0.0005       & 0.0000 +/- 0.0000                                               & 0.0000 +/- 0.0000                                                & 0.0000 +/- 0.0000                                                  & 0.4833 +/- 0.0005      & 0.0000 +/- 0.0000      \\ \cline{2-8}
                            & NB of Updates & 417415.25 +/- 289.20    & 417415.25 +/- 289.20                                            & 38825.60 +/- 541.45                                              & 8546.70 +/- 217.84                                                 & 417415.25 +/- 289.20   & 7086.10 +/- 213.04     \\ \cline{2-8}
                            & Cpu Time      & 19.7401 +/- 1.3051      & 21.5478 +/- 0.7875                                              & 20.0185 +/- 0.0683                                               & 22.3855 +/- 0.1676                                                 & 40.1595 +/- 1.0989     & 30.1619 +/- 1.9798     \\ \hline
\multirow{3}{*}{PA1}        & Mistake Rate  & 0.4833 +/- 0.0005       & 0.0000 +/- 0.0000                                               & 0.0000 +/- 0.0000                                                & 0.0000 +/- 0.0000                                                  & 0.0000 +/- 0.0000      & 0.0000 +/- 0.0000      \\ \cline{2-8}
                            & NB of Updates & 417415.25 +/- 289.20    & 165479.85 +/- 482.04                                            & 15909.95 +/- 362.64                                              & 7080.25 +/- 203.26                                                 & 6079.00 +/- 191.00     & 5858.20 +/- 171.80     \\ \cline{2-8}
                            & Cpu Time      & 18.3962 +/- 0.1999      & 20.2592 +/- 0.6342                                              & 20.2972 +/- 0.2150                                               & 24.0842 +/- 1.1345                                                 & 27.4099 +/- 1.3382     & 41.6198 +/- 1.5887     \\ \hline
\multirow{3}{*}{PA}         & Mistake Rate  & 0.4833 +/- 0.0005       & 0.0000 +/- 0.0000                                               & 0.0000 +/- 0.0000                                                & 0.0000 +/- 0.0000                                                  & 0.0000 +/- 0.0000      & 0.4833 +/- 0.0005      \\ \cline{2-8}
                            & NB of Updates & 417415.25 +/- 289.20    & 165479.85 +/- 482.04                                            & 15909.95 +/- 362.64                                              & 7080.25 +/- 203.26                                                 & 6079.00 +/- 191.00     & 417415.25 +/- 289.20   \\ \cline{2-8}
                            & Cpu Time      & 24.8938 +/- 0.3231      & 18.5542 +/- 0.0401                                              & 20.3547 +/- 0.0467                                               & 23.5226 +/- 0.0287                                                 & 30.2624 +/- 0.1127     & 64.7397 +/- 0.7773     \\ \hline
\multirow{3}{*}{OGD}        & Mistake Rate  & 0.4676 +/- 0.0004       & 0.0894 +/- 0.0003                                               & 0.0008 +/- 0.0000                                                & 0.0000 +/- 0.0000                                                  & 0.0000 +/- 0.0000      & 0.0000 +/- 0.0000      \\ \cline{2-8}
                            & NB of Updates & 271826.85 +/- 243.17    & 52014.10 +/- 164.87                                             & 485.65 +/- 20.00                                                 & 0.05 +/- 0.22                                                      & 0.00 +/- 0.00          & 0.00 +/- 0.00          \\ \cline{2-8}
                            & Cpu Time      & 18.1837 +/- 0.0900      & 19.1052 +/- 0.6240                                              & 21.2498 +/- 0.7872                                               & 25.5130 +/- 0.7948                                                 & 31.5751 +/- 0.0787     & 47.0398 +/- 0.4020     \\ \hline
\multirow{3}{*}{NHERD}      & Mistake Rate  & 0.2846 +/- 0.0148       & 0.2782 +/- 0.0140                                               & 0.2726 +/- 0.0133                                                & 0.2679 +/- 0.0126                                                  & 0.2636 +/- 0.0116      & 0.2597 +/- 0.0103      \\ \cline{2-8}
                            & NB of Updates & 501082.60 +/- 19966.44  & 505976.10 +/- 15737.72                                          & 510399.05 +/- 12360.74                                           & 514404.10 +/- 10132.99                                             & 518129.25 +/- 8674.91  & 521487.25 +/- 7336.42  \\ \cline{2-8}
                            & Cpu Time      & 28.4037 +/- 0.5968      & 37.0307 +/- 0.4370                                              & 49.3629 +/- 0.8756                                               & 78.1170 +/- 0.9857                                                 & 137.9415 +/- 6.6318    & 247.7766 +/- 1.8430    \\ \hline
\multirow{3}{*}{NAROW}      & Mistake Rate  & 0.3561 +/- 0.0244       & 0.3560 +/- 0.0244                                               & 0.3560 +/- 0.0244                                                & 0.3561 +/- 0.0244                                                  & 0.3561 +/- 0.0244      & 0.3561 +/- 0.0244      \\ \cline{2-8}
                            & NB of Updates & 547938.10 +/- 11540.63  & 547938.10 +/- 11540.63                                          & 547938.10 +/- 11540.63                                           & 547938.10 +/- 11540.63                                             & 547938.10 +/- 11540.63 & 547938.10 +/- 11540.63 \\ \cline{2-8}
                            & Cpu Time      & 30.8912 +/- 0.3900      & 39.7224 +/- 0.3541                                              & 61.5657 +/- 6.1638                                               & 110.4803 +/- 2.3765                                                & 168.1219 +/- 3.1345    & 325.0857 +/- 16.2145   \\ \hline
\multirow{3}{*}{CW}         & Mistake Rate  & 0.4479 +/- 0.0332       & 0.4479 +/- 0.033                                                & 0.4184 +/- 0.0346                                                & 0.4677 +/- 0.0344                                                  & 0.4626 +/- 0.0453      & 0.4623 +/- 0.0306      \\ \cline{2-8}
                            & NB of Updates & 169926.95 +/- 117124.46 & 169926.95 +/- 117124.46                                         & 106256.85 +/- 113484.30                                          & 84378.20 +/- 117066.68 & 80808.30 +/- 113258.68 & 78119.55 +/- 112694.68 \\ \cline{2-8}
                            & Cpu Time      & 33.4753 +/- 4.9887      & 33.6197 +/- 4.0935                                              & 40.6747 +/- 4.7547                                               & 55.3777 +/- 8.1420                                                 & 86.5521 +/- 14.1124    & 151.8714 +/- 26.4770   \\ \hline
\multirow{3}{*}{AROW}       & Mistake Rate  & 0.2436 +/- 0.0002       & 0.2433 +/- 0.0002                                               & 0.2432 +/- 0.0002                                                & 0.2432 +/- 0.0002                                                  & 0.2431 +/- 0.0002      & 0.2431 +/- 0.0002      \\ \cline{2-8}
                            & NB of Updates & 530783.00 +/- 457.05    & 530667.55 +/- 481.44                                            & 530652.60 +/- 468.13                                             & 530629.55 +/- 470.26                                               & 530628.30 +/- 462.17   & 530625.95 +/- 478.81   \\ \cline{2-8}
                            & Cpu Time      & 28.0119 +/- 0.0925      & 35.5127 +/- 0.0853                                              & 52.5572 +/- 2.0039                                               & 86.3058 +/- 6.6525                                                 & 138.2530 +/- 10.1840   & 289.4147 +/- 19.6682   \\ \hline
\multirow{3}{*}{SCW2}       & Mistake Rate  &     0.2389 +/- 0.0007                     &                                                                 0.2378 +/- 0.0002 &         0.2330 +/- 0.0002                             &                                                                    0.2312 +/- 0.0002 &   0.2295 +/- 0.0002                     &   0.2283 +/- 0.0002                      \\ \cline{2-8}
                            & NB of Updates &   458522.00 +/- 2659.43                       &                                                                532590.85 +/- 217.59 & 529978.80 +/- 206.18                                                                 &    530110.10 +/- 203.86                                                                &                        527576.40 +/- 213.41 &             524764.75 +/- 217.54             \\ \cline{2-8}
                            & Cpu Time      & 33.0266 +/- 1.1646                        &                                                                 47.8083 +/- 0.8593 &    60.5349 +/- 3.5711                                                              &99.0459 +/- 5.6878                                                                    &                       188.9169 +/- 1.4112 &   295.2806 +/- 11.2137                      \\ \hline
\multirow{3}{*}{ALMA}       & Mistake Rate  & 0.4839 +/- 0.0003       & 0.0000 +/- 0.0000                                               & 0.0000 +/- 0.0000                                                & 0.0000 +/- 0.0000                                                  & 0.0000 +/- 0.0000      & 0.0000 +/- 0.0000      \\ \cline{2-8}
                            & NB of Updates & 281153.20 +/- 192.06    & 0.00 +/- 0.00                                                   & 0.00 +/- 0.00                                                    & 0.00 +/-  0.00 & 0.00 +/- 0.00          & 0.00 +/- 0.00          \\ \cline{2-8}
                            & Cpu Time      & 19.6621 +/- 0.2444      & 20.9613 +/- 0.2967                                              & 20.9787 +/- 0.3217                                               & 24.0590 +/- 0.2602                                                 & 29.8072 +/- 0.7551     & 46.9257 +/- 3.1696     \\ \hline
\multirow{3}{*}{aROMMA}     & Mistake Rate  & 0.4839 +/- 0.0137       & 0.0000 +/- 0.0000                                               & 0.0000 +/- 0.0000                                                & 0.0000 +/- 0.0000                                                  & 0.0000 +/- 0.0000      & 0.0000 +/- 0.0000      \\ \cline{2-8}
                            & NB of Updates & 281167.85 +/- 7943.77   & 101516.55 +/- 6832.94                                           & 7313.35 +/- 3788.01                                              & 2756.40 +/- 1394.49                                                & 1914.50 +/- 1134.72    & 1606.50 +/- 1082.88    \\ \cline{2-8}
                            & Cpu Time      & 19.7463 +/- 1.4132      & 19.3363 +/- 0.1313                                              & 20.9656 +/- 0.4234                                               & 24.4997 +/- 0.1859                                                 & 33.7338 +/- 2.7316     & 47.7605 +/- 1.2837     \\ \hline
\multirow{3}{*}{IELLIP}     & Mistake Rate  & 0.4827 +/- 0.0005       & 0.0000 +/- 0.0000                                               & 0.0000 +/- 0.0000                                                & 0.0000 +/- 0.0000                                                  & 0.0000 +/- 0.0000      & 0.0000 +/- 0.0000      \\ \cline{2-8}
                            & NB of Updates & 280440.90 +/- 268.31    & 0.00 +/- 0.00                                                   & 0.00 +/- 0.00                                                    & 0.00 +/- 0.00                                                      & 0.00 +/- 0.00          & 0.00 +/- 0.00          \\ \cline{2-8}
                            & Cpu Time      & 29.4976 +/- 0.3621      & 31.7600 +/- 0.4450                                              & 36.4516 +/- 0.5889                                               & 44.2916 +/- 1.4691                                                 & 64.8225 +/- 4.2496     & 95.6598 +/- 2.4087     \\ \hline
\end{tabular}

\end{adjustbox}
\end{table}
\end{landscape}

\begin{landscape}
\begin{table}[ht]
\caption {Results for mnist multiclass dataset} \label{tab:tab5}
\begin{adjustbox}{width=1.5\textwidth}
\begin{tabular}{|l|l|l|l|l|l|l|l|}
\hline
\multicolumn{8}{|l|}{\textbf{Database name: mnist (n=60000,d=780,No. of classes=10) nb of runs (updates) : 20}}                                                                                                            \\ \hline
\textbf{Algorithm}                       &               & \textbf{m=1}                     & \textbf{m=2}                     & \textbf{m=4}                 & \textbf{m=8}                   & \textbf{m=16}                   & \textbf{m=32}                   \\ \hline
\multirow{3}{*}{M\_PA}          & Mistake Rate  & 0.1445 +/- 0.0007       & 0.0087 +/- 0.0003       & 0.0000 +/- 0.0000    & 0.0000 +/- 0.0000     & 0.0000 +/- 0.0000      & 0.0000 +/- 0.0000      \\ \cline{2-8}
                                & NB of Updates & 23340.50 +/- 105.42     & 14960.25 +/- 100.96     & 5179.90 +/- 75.45    & 2047.00 +/- 57.02     & 685.00 +/- 28.04       & 147.40 +/- 12.84       \\ \cline{2-8}
                                & Cpu Time      & 3.9079 +/- 0.0976       & 4.6027 +/- 0.0327       & 5.8063 +/- 0.0583    & 8.0101 +/- 0.0999     & 12.1759 +/- 0.5533     & 19.0469 +/- 0.0862     \\ \hline
\multirow{3}{*}{M\_PA1}         & Mistake Rate  & 0.1445 +/- 0.0007       & 0.0087 +/- 0.0003       & 0.0000 +/- 0.0000    & 0.0000 +/- 0.0000     & 0.0000 +/- 0.0000      & 0.0000 +/- 0.0000      \\ \cline{2-8}
                                & NB of Updates & 23340.50 +/- 105.42     & 14960.25 +/- 100.96     & 5179.90 +/- 75.45    & 2047.00 +/- 57.02     & 685.00 +/- 28.04       & 147.40 +/- 12.84       \\ \cline{2-8}
                                & Cpu Time      & 3.7608 +/- 0.0831       & 4.4899 +/- 0.0337       & 5.6725 +/- 0.0421    & 8.0563 +/- 0.2748     & 11.6835 +/- 0.0773     & 19.1395 +/- 0.0907     \\ \hline
\multirow{3}{*}{M\_PA2}         & Mistake Rate  & 0.1445 +/- 0.0007       & 0.0087 +/- 0.00003      & 0.0000 +/- 0.0000    & 0.0000 +/- 0.0000     & 0.0000 +/- 0.0000      & 0.0000 +/- 0.0000      \\ \cline{2-8}
                                & NB of Updates & 23340.50 +/- 105.42     & 22704.90 +/- 98.47      & 7928.90 +/- 86.60    & 2368.10 +/- 53.01     & 732.40 +/- 23.10       & 158.10 +/- 10.24       \\ \cline{2-8}
                                & Cpu Time      & 4.3315 +/- 0.0727       & 4.6526 +/- 0.0538       & 6.0112 +/- 0.0464    & 10.8198 +/- 0.9914    & 14.0190 +/- 1.9090     & 14.0190 +/- 1.9090     \\ \hline
\multirow{3}{*}{M\_OGD}         & Mistake Rate  & 0.1207 +/- 0.0015       & 0.0655 +/- 0.0016       & 0.0187 +/- 0.0013    & 0.0024 +/- 0.0003     & 0.0001 +/- 0.0000      & 0.0000 +/- 0.0000      \\ \cline{2-8}
                                & NB of Updates & 7246.20 +/- 88.40       & 3932.05 +/- 97.74       & 1119.75 +/- 79.36    & 141.40 +/- 18.50      & 5.75 +/- 1.77          & 0.05 +/- 0.22          \\ \cline{2-8}
                                & Cpu Time      & 3.5169 +/- 0.0414       & 4.2101 +/- 0.0610       & 5.5252 +/- 0.2823    & 7.6126 +/- 0.6102     & 11.0401+/-0.0708       & 18.9730 +/- 0.0990     \\ \hline
\multirow{3}{*}{M\_ROMMA}       & Mistake Rate  & 0.1893 +/- 0.0033       & 0.0410 +/- 0.0013       & 0.0019 +/- 0.0003    & 0.0000 +/- 0.0000     & 0.0000 +/- 0.0000      & 0.0000 +/- 0.0000      \\ \cline{2-8}
                                & NB of Updates & 11358.50 +/- 200.15     & 2462.85 +/- 79.01       & 114.95 +/- 16.94     & 0.05 +/- 0.22         & 0.00 +/- 0.00          & 0.00 +/- 0.00          \\ \cline{2-8}
                                & Cpu Time      & 3.9515 +/- 0.0584       & 4.5683 +/- 0.0434       & 5.6035 +/- 0.0354    & 8.0510 +/- 0.7452     & 11.0940 +/- 0.1235     & 18.4482 +/- 0.1118     \\ \hline
\multirow{3}{*}{M\_PerceptronM} & Mistake Rate  & 0.1538 +/- 0.0010       & 0.0232+/-0.0003         & 0.0010+/-0.0001      & 0.0000 +/- 0.0000     & 0.0000 +/- 0.0000      & 0.0000 +/- 0.0000      \\ \cline{2-8}
                                & NB of Updates & 9226.00 +/- 61.78       & 1393.15 +/- 20.73       & 59.25 +/- 5.43       & 0.80 +/- 0.77         & 0.00 +/- 0.00          & 0.00 +/- 0.00          \\ \cline{2-8}
                                & Cpu Time      & 3.4645 +/- 0.2610       & 4.0306 +/- 0.0402       & 4.9131 +/- 0.0361    & 6.7701 +/- 0.0197     & 10.6000 +/- 0.0303     & 18.1062 +/- 0.0280     \\ \hline
\multirow{3}{*}{M\_PerceptronS} & Mistake Rate  & 0.1473 +/- 0.0008       & 0.0222 +/- 0.0004       & 0.0012 +/- 0.0001    & 0.0000 +/- 0.0000     & 0.0000 +/- 0.0000      & 0.0000 +/- 0.0000      \\ \cline{2-8}
                                & NB of Updates & 8836.95 +/- 47.71       & 1332.55 +/- 23.93       & 70.50 +/- 5.36       & 1.25 +/- 0.97         & 0.00 +/- 0.00          & 0.00 +/- 0.00          \\ \cline{2-8}
                                & Cpu Time      & 3.2463 +/- 0.0270       & 3.7111 +/- 0.0213       & 4.5197 +/- 0.0402    & 5.9983 +/- 0.0543     & 8.7511 +/- 0.0262      & 14.5107 +/- 0.0312     \\ \hline
\multirow{3}{*}{M\_PerceptronU} & Mistake Rate  & 0.1428 +/- 0.0008       & 0.0287 +/- 0.0005       & 0.0022 +/- 0.0001    & 0.0001 +/- 0.0000     & 0.0000 +/- 0.0000      & 0.0000 +/- 0.0000      \\ \cline{2-8}
                                & NB of Updates & 8569.65 +/- 49.48       & 1721.85 +/- 29.30       & 130.35 +/- 7.92      & 3.50 +/- 1.10         & 0.00 +/- 0.00          & 0.00 +/- 0.00          \\ \cline{2-8}
                                & Cpu Time      & 3.2187 +/- 0.0145       & 3.6523 +/- 0.0109       & 4.4571 +/- 0.0451    & 6.2330 +/- 0.3464     & 10.1010 +/- 1.6098     & 14.0720 +/- 0.0741     \\ \hline
\multirow{3}{*}{M\_SCW2}        & Mistake Rate  & 0.1300 +/- 0.0010       & 0.0083 +/- 0.0003       & 0.0000 +/- 0.0000    & 0.0000 +/- 0.0000     & 0.0000 +/- 0.0000      & 0.0000 +/- 0.0000      \\ \cline{2-8}
                                & NB of Updates & 15185.10 +/- 89.50      & 3110.40 +/- 47.77       & 6.10 +/- 2.29        & 0.05 +/- 0.22         & 0.05 +/- 0.22          & 0.05 +/- 0.22          \\ \cline{2-8}
                                & Cpu Time      & 1209.3032 +/- 3273.0507 & 638.4680 +/- 3.4593     & 573.3860 +/- 3.0892  & 402.3738 +/- 1.9610   & 449.8596 +/- 20.2444   & 483.6398 +/- 2.6823    \\ \hline
\multirow{3}{*}{M\_SCW1}        & Mistake Rate  & 0.1902 +/- 0.0040       & 0.1337 +/- 0.0056       & 0.1298 +/- 0.0066    & 0.1301 +/- 0.0067     & 0.1304 +/- 0.0067      & 0.1305 +/- 0.0067      \\ \cline{2-8}
                                & NB of Updates & 13900.25 +/- 224.84     & 8752.30 +/- 329.94      & 7789.65 +/- 397.70   & 7810.20 +/- 404.02    & 7823.05 +/- 404.57     & 7830.25 +/- 404.16     \\ \cline{2-8}
                                & Cpu Time      & 340.8704 +/- 5.1127     & 3128.8948 +/- 8758.0710 & 951.4627 +/- 36.5136 & 1620.4285 +/- 88.5287 & 3117.9875 +/- 152.6886 & 6421.7155 +/- 364.9498 \\ \hline
\multirow{3}{*}{M\_CW}          & Mistake Rate  & 0.1328 +/- 0.0009       & 0.0081 +/- 0.0003       & 0.0000 +/- 0.0000    & 0.0000 +/- 0.0000     & 0.0000 +/- 0.0000      & 0.0000 +/- 0.0000      \\ \cline{2-8}
                                & NB of Updates & 15309.60 +/- 85.36      & 3090.45 +/- 49.21       & 3.85 +/- 1.18        & 0.00 +/- 0.00         & 0.00 +/- 0.00          & 0.00 +/- 0.00          \\ \cline{2-8}
                                & Cpu Time      & 363.4854 +/- 4.3348     & 442.6689 +/- 15.7587    & 421.5143 +/- 2.3571  & 415.2526 +/- 8.0262   & 664.9018 +/- 73.7706   & 630.8751 +/- 5.3968    \\ \hline

\multirow{3}{*}{M\_aROMMA}          & Mistake Rate  & 0.1893 +/- 0.0033       & 0.0410 +/- 0.0013       & 0.0019 +/- 0.0003    & 0.0000 +/- 0.0000     & 0.0000 +/- 0.0000      & 0.0000 +/- 0.0000      \\ \cline{2-8}
                                & NB of Updates & 11360.60 +/- 200.60      & 6811.80 +/- 144.72      & 2435.65 +/- 218.52        & 901.10 +/- 201.43         & 276.85 +/- 107.88        & 69.45 +/- 44.28          \\ \cline{2-8}
                                & Cpu Time      & 5.1332 +/- 0.1563     & 7.1778 +/- 0.0635    & 11.3280 +/- 0.0858  & 19.5403 +/- 0.1616   & 36.7845 +/- 0.5950   & 70.3360 +/- 1.2428   \\ \hline
\multirow{3}{*}{M\_AROW}          & Mistake Rate  & 0.3783 +/- 0.0477       & 0.3272 +/- 0.0454       & 0.3044 +/- 0.0515    & 0.2971 +/- 0.0616     & 0.3296 +/- 0.0661 & 0.3148 +/- 0.0023      \\ \cline{2-8}
                                & NB of Updates & 26633.75 +/- 2087.69      & 24248.55 +/- 1910.37      &  23424.70 +/- 2092.56       & 23416.45 +/- 2865.54         & 24240.05 +/- 3194.06        & 23121.50 +/- 383.96         \\ \cline{2-8}
                                & Cpu Time      & 967.2876 +/- 106.1994    & 1559.1950 +/- 128.7375    & 2063.1689 +/- 177.9005  & 4890.8882 +/- 562.3463   & 8203.4645 +/- 1054.7809  & 19121.6330 +/- 234.8003  \\ \hline

\end{tabular}

\end{adjustbox}

\end{table}
\end{landscape}
\begin{landscape}
\begin{table}[ht]
\caption {Results for glass multiclass dataset} \label{tab:tab6}
\begin{adjustbox}{width=1.5\textwidth}
\begin{tabular}{|l|l|l|l|l|l|l|l|}
\hline
\multicolumn{8}{|l|}{\textbf{Database name: glass (n=214,d=300,No. of classes=6)   nb of runs (updates):20} }                                                                                  \\ \hline
\textbf{Algorithm}                       &               & \textbf{m=1}              & \textbf{m=2}               & \textbf{m=4}               & \textbf{m=8}               & \textbf{m=16 }             & \textbf{m=32}               \\ \hline
\multirow{3}{*}{M\_PA}          & Mistake Rate  & 0.5832 +/- 0.0326 & 0.0049 +/- 0.0035 & 0.0000 +/- 0.0000 & 0.0000 +/- 0.0000 & 0.0000 +/- 0.0000 & 0.0000 +/- 0.0000  \\ \cline{2-8}
                                & NB of Updates & 189.50 +/- 5.89   & 127.80 +/- 7.52   & 26.30 +/- 6.67    & 3.00 +/- 1.78     & 0.90 +/- 1.07     & 0.55 +/- 0.76      \\ \cline{2-8}
                                & Cpu Time      & 0.0103 +/- 0.0005 & 0.0124 +/- 0.0004 & 0.0154 +/- 0.0003 & 0.0202 +/- 0.0003 & 0.0312 +/- 0.0017 & 0.0510 +/- 0.0011  \\ \hline
\multirow{3}{*}{M\_PA1}         & Mistake Rate  & 0.5605 +/- 0.0150 & 0.0049 +/- 0.0035 & 0.0000 +/- 0.0000 & 0.0000 +/- 0.0000 & 0.0000 +/- 0.0000 & 0.0000 +/- 0.0000  \\ \cline{2-8}
                                & NB of Updates & 206.75 +/- 2.61   & 127.80 +/- 7.52   & 26.30 +/- 6.67    & 3.00 +/- 1.78     & 0.90 +/- 1.07     & 0.55 +/- 0.76      \\ \cline{2-8}
                                & Cpu Time      & 0.0107 +/- 0.0004 & 0.0125 +/- 0.0003 & 0.0154 +/- 0.0004 & 0.0204 +/- 0.0003 & 0.0304 +/- 0.0005 & 0.0509 +/- 0.0011  \\ \hline
\multirow{3}{*}{M\_PA2}         & Mistake Rate  & 0.5619 +/- 0.0271 & 0.0049 +/- 0.0035 & 0.0000 +/- 0.0000 & 0.0000 +/- 0.0000 & 0.0000 +/- 0.0000 & 0.0000 +/- 0.0000  \\ \cline{2-8}
                                & NB of Updates & 212.30 +/- 1.38   & 181.60 +/- 6.92   & 180.95 +/- 7.61   & 131.65 +/- 8.65   & 1.85 +/- 1.14     & 0.45 +/- 0.51      \\ \cline{2-8}
                                & Cpu Time      & 0.0107 +/- 0.0005 & 0.0129 +/- 0.0006 & 0.0178 +/- 0.0003 & 0.0277 +/- 0.0007 & 0.0385 +/- 0.0008 & 0.0588 +/- 0.0040  \\ \hline
\multirow{3}{*}{M\_OGD}         & Mistake Rate  & 0.5533 +/- 0.0313 & 0.1610 +/- 0.0222 & 0.0035 +/- 0.0048 & 0.0000 +/- 0.0000 & 0.0000 +/- 0.0000 & 0.0000 +/- 0.0000  \\ \cline{2-8}
                                & NB of Updates & 165.85 +/- 6.58   & 49.80 +/- 5.31    & 0.85 +/- 1.09     & 0.00 +/- 0.00     & 0.00 +/- 0.00     & 0.00 +/- 0.00      \\ \cline{2-8}
                                & Cpu Time      & 0.0102 +/- 0.0004 & 0.0112 +/- 0.0004 & 0.0137 +/- 0.0003 & 0.0187 +/- 0.0003 & 0.0293 +/- 0.0006 & 0.0500 +/- 0.0009  \\ \hline
\multirow{3}{*}{M\_ROMMA}       & Mistake Rate  & 0.5963 +/- 0.0493 & 0.0640 +/- 0.0426 & 0.0002 +/- 0.0010 & 0.0000 +/- 0.0000 & 0.0000 +/- 0.0000 & 0.0000 +/- 0.0000  \\ \cline{2-8}
                                & NB of Updates & 127.60 +/- 10.54  & 13.70 +/- 9.11    & 0.05 +/- 0.22     & 0.00 +/- 0.00     & 0.00 +/- 0.00     & 0.00 +/- 0.00      \\ \cline{2-8}
                                & Cpu Time      & 0.0108 +/- 0.0030 & 0.0117 +/- 0.0026 & 0.0145 +/- 0.0033 & 0.0198 +/- 0.0030 & 0.0290 +/- 0.0029 & 0.0495 +/- 0.0035  \\ \hline
\multirow{3}{*}{M\_PerceptronM} & Mistake Rate  & 0.5967 +/- 0.0274 & .0582 +/- 0.0171  & 0.0000 +/- 0.0000 & 0.0000 +/- 0.0000 & 0.0000 +/- 0.0000 & 0.0000 +/- 0.0000  \\ \cline{2-8}
                                & NB of Updates & 127.70 +/- 5.86   & 12.45 +/- 3.66    & 0.00 +/- 0.00     & 0.00 +/- 0.00     & 0.00 +/- 0.00     & 0.00 +/- 0.00      \\ \cline{2-8}
                                & Cpu Time      & 0.0100 +/- 0.0019 & 0.0110 +/- 0.0022 & 0.0134 +/- 0.0022 & 0.0183 +/- 0.0015 & 0.0295 +/- 0.0039 & 0.0480 +/- 0.0028  \\ \hline
\multirow{3}{*}{M\_PerceptronS} & Mistake Rate  & 0.5780 +/- 0.0317 & 0.0586 +/- 0.0164 & 0.0000 +/- 0.0000 & 0.0000 +/- 0.0000 & 0.0000 +/- 0.0000 & 0.0000 +/- 0.0000  \\ \cline{2-8}
                                & NB of Updates & 123.70 +/- 6.79   & 12.55 +/- 3.52    & 0.00 +/- 0.00     & 0.00 +/- 0.00     & 0.00 +/- 0.00     & 0.00 +/- 0.00      \\ \cline{2-8}
                                & Cpu Time      & 0.0117 +/- 0.0077 & 0.0115 +/- 0.0038 & 0.0132 +/- 0.0037 & 0.0168 +/- 0.0044 & 0.0233 +/- 0.0045 & 0.0365 +/- 0.0039  \\ \hline
\multirow{3}{*}{M\_PerceptronU} & Mistake Rate  & 0.5682 +/- 0.0308 & 0.1269 +/- 0.0246 & 0.0000 +/- 0.0000 & 0.0000 +/- 0.0000 & 0.0000 +/- 0.0000 & 0.0000 +/- 0.0000  \\ \cline{2-8}
                                & NB of Updates & 121.60 +/- 6.59   & 27.15 +/- 5.25    & 0.00 +/- 0.00     & 0.00 +/- 0.00     & 0.00 +/- 0.00     & 0.00 +/- 0.00      \\ \cline{2-8}
                                & Cpu Time      & 0.0101 +/- 0.0026 & 0.0113 +/- 0.0035 & 0.0131 +/- 0.0037 & 0.0166 +/- 0.0038 & 0.0235 +/- 0.0039 & 0.0363 +/- 0.0041  \\ \hline
\multirow{3}{*}{M\_SCW2}        & Mistake Rate  & 0.4832 +/- 0.0251 & 0.0079 +/- 0.0046 & 0.0000 +/- 0.0000 & 0.0000 +/- 0.0000 & 0.0002 +/- 0.0010 & 0.0000 +/- 0.0000  \\ \cline{2-8}
                                & NB of Updates & 180.80 +/- 7.35   & 47.50 +/- 5.49    & 1.55 +/- 1.15     & 0.00 +/- 0.00     & 0.65 +/- 0.99     & 0.10 +/- 0.31      \\ \cline{2-8}
                                & Cpu Time      & 0.0146 +/- 0.0004 & 0.0182 +/- 0.0023 & 0.0224 +/- 0.0008 & 0.0306 +/- 0.0005 & 0.0490 +/- 0.0016 & 0.0809 +/- 0.0040  \\ \hline
\multirow{3}{*}{M\_SCW1}        & Mistake Rate  & 0.4722 +/- 0.0231 & 0.2668 +/- 0.0220 & 0.1769 +/- 0.0186 & 0.1203 +/- 0.0176 & 0.0815 +/- 0.0237 & 0.0495 +/- 0.0200  \\ \cline{2-8}
                                & NB of Updates & 157.65 +/- 5.44   & 106.05 +/- 8.39   & 65.00 +/- 6.77    & 41.55 +/- 5.66    & 26.40 +/- 4.97    & 15.80 +/- 5.42     \\ \cline{2-8}
                                & Cpu Time      & 0.0147 +/- 0.0039 & 0.0193 +/- 0.0054 & 0.0264 +/- 0.0045 & 0.0378 +/- 0.0048 & 0.0603 +/- 0.0045 & 0.0988 +/- 0.0067  \\ \hline
\multirow{3}{*}{M\_CW}          & Mistake Rate  & 0.5229 +/- 0.0311 & 0.0086 +/- 0.0055 & 0.0000 +/- 0.0000 & 0.0000 +/- 0.0000 & 0.0000 +/- 0.0000 & 0.0000 +/- 0.0000  \\ \cline{2-8}
                                & NB of Updates & 157.25 +/- 7.43   & 46.40 +/- 5.29    & 0.00 +/- 0.00     & 0.00 +/- 0.00     & 0.00 +/- 0.00     & 0.00 +/- 0.00      \\ \cline{2-8}
                                & Cpu Time      & 0.0133 +/- 0.0005 & 0.0159 +/- 0.0003 & 0.0193 +/- 0.0005 & 0.0274 +/- 0.0007 & 0.0442 +/- 0.0017 & 0.0747 +/- 0.0018  \\ \hline
\multirow{3}{*}{M\_AROW}        & Mistake Rate  & 0.4918 +/- 0.0251 & 0.3266 +/- 0.0327 & 0.2883 +/- 0.0301 & 0.2612 +/- 0.0160 & 0.2554 +/- 0.0180 & 0.2505 +/- 0.0178  \\ \cline{2-8}
                                & NB of Updates & 213.85 +/- 0.37   & 187.35 +/- 4.23   & 188.45 +/- 4.33   & 207.75 +/- 2.38   & 207.90 +/- 2.49   & 207.85 +/- 2.46    \\ \cline{2-8}
                                & Cpu Time      & 0.0144 +/- 0.0004 & 0.0192 +/- 0.0004 & 0.0301 +/- 0.0007 & 0.0536 +/- 0.0014 & 0.0971 +/- 0.0018 & 0.1812 +/- 0.0018  \\ \hline
\multirow{3}{*}{M\_aROMMA}      & Mistake Rate  & 0.5888 +/- 0.0395 & 0.0584 +/- 0.0372 & 0.0002 +/- 0.0010 & 0.0000 +/- 0.0000 & 0.0000 +/- 0.0000 & 0.0000 +/- 0.00000 \\ \cline{2-8}
                                & NB of Updates & 167.05 +/- 10.98  & 106.60 +/- 9.81   & 36.00 +/- 8.55    & 13.20 +/- 8.95    & 7.35 +/- 5.25     & 4.60 +/- 3.62      \\ \cline{2-8}
                                & Cpu Time      & 0.0115 +/- 0.0030 & 0.0139 +/- 0.0034 & 0.0173 +/- 0.0031 & 0.0250 +/- 0.0048 & 0.0362 +/- 0.0035 & 0.0595 +/- 0.0038  \\ \hline

\end{tabular}

\end{adjustbox}
\end{table}
\end{landscape}
\begin{landscape}
\begin{table}[ht]
\caption {Results for segment multiclass dataset} \label{tab:tab7}
\begin{adjustbox}{width=1.5\textwidth}
\begin{tabular}{|l|l|l|l|l|l|l|l|}
\hline
\multicolumn{8}{|l|}{Database name: segment (n=2310,d=19,No. of classes=7)          nb of runs (updates): 20}                                                                         \\ \hline
Algorithm                       &               & m=1                & m=2               & m=4               & m=8               & m=16              & m=32              \\ \hline
\multirow{3}{*}{M\_PA}          & Mistake Rate  & 0.2103 +/- 0.0071  & 0.0037 +/- 0.0011 & 0.0000 +/- 0.0000 & 0.0000 +/- 0.0000 & 0.0000 +/- 0.0000 & 0.0000 +/- 0.0000 \\ \cline{2-8}
                                & NB of Updates & 1372.00 +/- 23.45  & 805.35 +/- 16.12  & 258.10 +/- 12.43  & 90.50 +/- 15.51   & 36.15 +/- 9.96    & 12.90 +/- 6.49    \\ \cline{2-8}
                                & Cpu Time      & 0.0954 +/- 0.0048  & 0.1119 +/- 0.0011 & 0.1438 +/- 0.0053 & 0.2025 +/- 0.0036 & 0.3107 +/- 0.0100 & 0.5276 +/- 0.0103 \\ \hline
\multirow{3}{*}{M\_PA1}         & Mistake Rate  & 0.1993 +/- 0.0066  & 0.0037 +/- 0.0011 & 0.0000 +/- 0.0000 & 0.0000 +/- 0.0000 & 0.0000 +/- 0.0000 & 0.0000 +/- 0.0000 \\ \cline{2-8}
                                & NB of Updates & 1419.80 +/- 15.88  & 805.35 +/- 16.12  & 258.10 +/- 12.43  & 90.50 +/- 15.51   & 36.15 +/- 9.96    & 12.90 +/- 6.49    \\ \cline{2-8}
                                & Cpu Time      & 0.0963 +/- 0.0020  & 0.1129 +/- 0.0012 & 0.1449 +/- 0.0068 & 0.2025 +/- 0.0060 & 0.3190 +/- 0.0092 & 0.5219 +/- 0.0050 \\ \hline
\multirow{3}{*}{M\_PA2}         & Mistake Rate  & 0.2034 +/- 0.0074  & 0.0038 +/- 0.0010 & 0.0000 +/- 0.0000 & 0.0000 +/- 0.0000 & 0.0000 +/- 0.0000 & 0.0000 +/- 0.0000 \\ \cline{2-8}
                                & NB of Updates & 1620.50 +/- 17.84  & 1312.50 +/- 20.10 & 1309.60 +/- 20.30 & 443.80 +/- 23.93  & 66.80 +/- 14.78   & 21.75 +/- 8.33    \\ \cline{2-8}
                                & Cpu Time      & 0.0981 +/- 0.0022  & 0.1177 +/- 0.0041 & 0.1658 +/- 0.0057 & 0.2403 +/- 0.0021 & 0.3552 +/- 0.0037 & 0.5766 +/- 0.0118 \\ \hline
\multirow{3}{*}{M\_OGD}         & Mistake Rate  & 0.1608 +/- 0.0062  & 0.0384 +/- 0.0054 & 0.0018 +/- 0.0011 & 0.0000 +/- 0.0001 & 0.0000 +/- 0.0000 & 0.0000 +/- 0.0000 \\ \cline{2-8}
                                & NB of Updates & 964.20 +/- 25.42   & 104.55 +/- 13.95  & 5.25 +/- 3.18     & 0.05 +/- 0.22     & 0.00 +/- 0.00     & 0.00 +/- 0.00     \\ \cline{2-8}
                                & Cpu Time      & 0.0910 +/- 0.0022  & 0.0989 +/- 0.0015 & 0.1255 +/- 0.0010 & 0.1841 +/- 0.0082 & 0.2869 +/- 0.0025 & 0.4995 +/- 0.0047 \\ \hline
\multirow{3}{*}{M\_ROMMA}       & Mistake Rate  & 0.2182 +/- 0.0079  & 0.0082 +/- 0.0034 & 0.0001 +/- 0.0002 & 0.0000 +/- 0.0000 & 0.0000 +/- 0.0000 & 0.0000 +/- 0.0000 \\ \cline{2-8}
                                & NB of Updates & 503.95 +/- 18.18   & 19.00 +/- 7.77    & 0.25 +/- 0.44     & 0.00 +/- 0.00     & 0.00 +/- 0.00     & 0.00 +/- 0.00     \\ \cline{2-8}
                                & Cpu Time      & 0.0881 +/- 0.0075  & 0.1004 +/- 0.0039 & 0.1269 +/- 0.0038 & 0.1811 +/- 0.0056 & 0.2824 +/- 0.0079 & 0.5001 +/- 0.0148 \\ \hline
\multirow{3}{*}{M\_PerceptronM} & Mistake Rate  & 0.2489 +/- 0.0058  & 0.0166 +/- 0.0021 & 0.0000 +/- 0.0001 & 0.0000 +/- 0.0000 & 0.0000 +/- 0.0000 & 0.0000 +/- 0.0000 \\ \cline{2-8}
                                & NB of Updates & 574.90 +/- 13.38   & 38.45 +/- 4.77    & 0.05 +/- 0.22     & 0.00 +/- 0.00     & 0.00 +/- 0.00     & 0.00 +/- 0.00     \\ \cline{2-8}
                                & Cpu Time      & 0.0848 +/- 0.0027  & 0.0967 +/- 0.0025 & 0.1250 +/- 0.0035 & 0.1760 +/- 0.0033 & 0.2796 +/- 0.0087 & 0.4838 +/- 0.0128 \\ \hline
\multirow{3}{*}{M\_PerceptronS} & Mistake Rate  & 0.2418 +/- 0.0059  & 0.0144 +/- 0.0015 & 0.0000 +/- 0.0001 & 0.0000 +/- 0.0000 & 0.0000 +/- 0.0000 & 0.0000 +/- 0.0000 \\ \cline{2-8}
                                & NB of Updates & 558.45 +/- 13.73   & 33.25 +/- 3.37    & 0.05 +/- 0.22     & 0.00 +/- 0.00     & 0.00 +/- 0.00     & 0.00 +/- 0.00     \\ \cline{2-8}
                                & Cpu Time      & 0.0844 +/- 0.0047  & 0.0940 +/- 0.0043 & 0.1161 +/- 0.0060 & 0.1511 +/- 0.0078 & 0.2259 +/- 0.0078 & 0.3737 +/- 0.0077 \\ \hline
\multirow{3}{*}{M\_PerceptronU} & Mistake Rate  & 0.2414 +/- 0.0063  & 0.0247 +/- 0.0030 & 0.0000 +/- 0.0001 & 0.0000 +/- 0.0000 & 0.0000 +/- 0.0000 & 0.0000 +/- 0.0000 \\ \cline{2-8}
                                & NB of Updates & 557.60 +/- 14.60   & 57.15 +/- 6.90    & 0.05 +/- 0.22     & 0.00 +/- 0.00     & 0.00 +/- 0.00     & 0.00 +/- 0.00     \\ \cline{2-8}
                                & Cpu Time      & 0.0807 +/- 0.0007  & 0.0932 +/- 0.0040 & 0.1099 +/- 0.0034 & 0.1494 +/- 0.0051 & 0.2235 +/- 0.0058 & 0.3565 +/- 0.0047 \\ \hline
\multirow{3}{*}{M\_SCW2}        & Mistake Rate  & 0.0915 +/- 0.0041  & 0.0032 +/- 0.0011 & 0.0000 +/- 0.0000 & 0.0000 +/- 0.0000 & 0.0000 +/- 0.0000 & 0.0000 +/- 0.0000 \\ \cline{2-8}
                                & NB of Updates & 578.50 +/- 14.16   & 116.50 +/- 7.70   & 0.90 +/- 0.72     & 0.00 +/- 0.00     & 0.00 +/- 0.00     & 0.00 +/- 0.00     \\ \cline{2-8}
                                & Cpu Time      & 0.1087 +/- 0.0053  & 0.1364 +/- 0.0013 & 0.1789 +/- 0.0051 & 0.2620 +/- 0.0043 & 0.4321 +/- 0.0056 & 0.7751 +/- 0.0058 \\ \hline
\multirow{3}{*}{M\_SCW1}        & Mistake Rate  & 0.0858 +/- 0.0041  & 0.0507 +/- 0.0029 & 0.0376 +/- 0.0031 & 0.0295 +/- 0.0035 & 0.0245 +/- 0.0036 & 0.0224 +/- 0.0042 \\ \cline{2-8}
                                & NB of Updates & 405.95 +/- 13.48   & 220.75 +/- 9.41   & 142.75 +/- 8.66   & 104.45 +/- 8.67   & 79.50 +/- 8.83    & 68.25 +/- 10.47   \\ \cline{2-8}
                                & Cpu Time      & 0.1064 +/- 0.0056  & 0.1363 +/- 0.0069 & 0.1854 +/- 0.0054 & 0.2873 +/- 0.0082 & 0.4849 +/- 0.0155 & 0.8574 +/- 0.0116 \\ \hline
\multirow{3}{*}{M\_CW}          & Mistake Rate  & 0.1187 +/- 0.0058  & 0.0042 +/- 0.0016 & 0.0000 +/- 0.0000 & 0.0000 +/- 0.0000 & 0.0000 +/- 0.0000 & 0.0000 +/- 0.0000 \\ \cline{2-8}
                                & NB of Updates & 534.30 +/- 13.67   & 64.75 +/- 6.04    & 0.00 +/- 0.00     & 0.00 +/- 0.00     & 0.00 +/- 0.00     & 0.00 +/- 0.00     \\ \cline{2-8}
                                & Cpu Time      & 0.1102 +/- 0.0047  & 0.1294 +/- 0.0013 & 0.1737 +/- 0.0025 & 0.2655 +/- 0.0105 & 0.4312 +/- 0.0066 & 0.7669 +/- 0.0074 \\ \hline
\multirow{3}{*}{M\_AROW}        & Mistake Rate  & 0.1089 +/- 0.0098  & 0.0922 +/- 0.0095 & 0.0867 +/- 0.0083 & 0.0841 +/- 0.0084 & 0.0841 +/- 0.0077 & 0.0949 +/- 0.0069 \\ \cline{2-8}
                                & NB of Updates & 1457.35 +/- 126.00 & 1548.45 +/- 84.19 & 1522.75 +/- 73.88 & 1488.70 +/- 72.76 & 1466.35 +/- 77.58 & 1839.15 +/- 42.89 \\ \cline{2-8}
                                & Cpu Time      & 0.1302 +/- 0.0045  & 0.1852 +/- 0.0044 & 0.2900 +/- 0.0131 & 0.4922 +/- 0.0121 & 0.8897 +/- 0.0292 & 1.9375 +/- 0.0334 \\ \hline
\multirow{3}{*}{M\_aROMMA}      & Mistake Rate  & 0.1931 +/- 0.0086  & 0.0088 +/- 0.0029 & 0.0001 +/- 0.0002 & 0.0000 +/- 0.0000 & 0.0000 +/- 0.0000 & 0.0000 +/- 0.0000 \\ \cline{2-8}
                                & NB of Updates & 1155.55 +/- 62.62  & 647.45 +/- 50.22  & 226.15 +/- 32.96  & 86.45 +/- 28.50   & 33.45 +/- 15.26   & 13.45 +/- 8.86    \\ \cline{2-8}
                                & Cpu Time      & 0.0985 +/- 0.0030  & 0.1220 +/- 0.0048 & 0.1575 +/- 0.0095 & 0.2323 +/- 0.0091 & 0.3630 +/- 0.0075 & 0.6277 +/- 0.0101 \\ \hline
\end{tabular}

\end{adjustbox}
\end{table}
\end{landscape}
\subsection{Comparison}
The $m=1$ value refer to working of original algorithms. We have updated weights for $m=2,4,8,16,32$ times and noticed that most of the algorithms achieved mistake rate close to zero. The convergence rate of representative algorithm are explained in their respective references. The limitation of MTWU is additional compilation time but we witnessed that zero mistake rate come for most of algorithms at $m \leq$ 4. This further strengthen the role of MTWU in online learning.\\

The online learning is real time prediction in data but lack in mistake rate as compare to batch processing. Using MTWU, it could overcome mistake rate challenge. The overburden extra time is very less using MTWU as each instance trained for very few iterations and most of algorithms achieve zero mistake rate for m $\leq$ 4 iterations. This study shows MTWU is an effective technique that has promising results to deliver. In particular, its use to multiple classes is praiseworthy as complexity to classify data increases with multiple classes. We have witnessed the importance of MTWU to both first and second order online learning algorithms. The disadvantage of MTWU is extra cost of running time but achieving zero mistake rate do not discourage the importance of MTWU. From this situation we feel that MTWU will be a useful to all platfrom of online learning algorithms to meet the real life data challenges.
\section{Concluding remarks and future directions}
In the present work, we have presented a novel approach to minimize mistake rate in online learning methods. Certainly, the state-of-art of the online learning algorithmis that it learn the model in online environments quickly and better regret bound. Also, mistake rate control is equally important. That is the reason why proposed technique MTWU is applicable in online learning to reduce mistake rate. The MTWU technique re-train the weights in online environments and for the single instance at a time. The validity of these technique have been provedwith different state-of-art algorithms. The experimentalresults observe that the proposed technique attains consistent and reliable results in different algorithms and datasets. The present research work examinesthe following imperative outcomes:\\
\begin{itemize}
\item The available work for online learning control mistake rate for single iteration only, but present work further minimize mistake rate with multiple iterations, and using small number of iterations.
\item The proposed research represents one of the first attempts in this direction.
\item The present study presents a significant analysis of different algorithms and datasets using proposed technique MTWU.
\item For justifying the proposed technique, the present work has been verified with more than twelve state-of-art algorithms and five benchmarked datasets including both binary and multi classes datasets.
\item The proposed technique is suitable to future algorithms in online learning.
\item The MTWU is very useful for reducing mistake rate of classification in large datasets with multiple classes.
\item The consistent experimental outcomes presented in the proposed study are without hugepreprocessing and it results in less time complexity.
\item The time complexity with MTWU for more than one iteration is not too expensive as compared to one iteration, and this strengthen propose technique.
\end{itemize}
Although, the online learning with MTWU needs attention of more researchers across the globe and its implementation inreal life scenarios requires rigorous experimentation, the present study is a breakthrough foronline learning.The future work also comprises the extension of present work to other big datasets and reducing both mistake rate and time cost. The MTWU technique could open scope to new online learning methods in future.




\end{document}